\documentclass[runningheads]{llncs}
\usepackage[T1]{fontenc}
\usepackage{graphicx}
\usepackage{booktabs}
\usepackage[misc]{ifsym}

\usepackage{mwe}
\usepackage{float}%提供float浮动环境
\usepackage{booktabs}%提供命令\toprule、\midrule、\bottomrule
\usepackage{multirow}%提供跨行命令\multirow{}{}{}
\usepackage{tabu}
\usepackage{hyperref}
\usepackage{url}
\usepackage{svg}
\usepackage{wrapfig}
\usepackage {subcaption}

\usepackage{times}
\usepackage{amsmath}
\usepackage{bm}
\usepackage{xcolor}
\usepackage{graphicx}
\usepackage[ruled,vlined]{algorithm2e}
\usepackage{paralist, tabularx}
\usepackage{enumitem}
\usepackage{amsfonts}
\usepackage{amssymb}

\newcommand{\xzb}[1]{\textcolor{black}{#1}}

\newcommand\zty[1]{{\color{black}#1}}
\newcommand\wy[1]{{\color{black}#1}}
\newcommand\lsy[1]{{\color{black}#1}}

\AtBeginDocument{%
  }
\begin{document}
\title{Simple Graph Condensation}

\toctitle{Simple Graph Condensation}
\tocauthor{Zhenbang Xiao, Yu Wang, Shunyu Liu, Huiqiong Wang, Mingli Song, Tongya Zheng}

\author{Zhenbang Xiao\inst{1} \and
Yu Wang\inst{1} \and
Shunyu Liu\inst{1} \and
Huiqiong Wang\inst{1}\textsuperscript{\Letter} \and
Mingli Song\inst{1} \and
Tongya Zheng\inst{1,2}}
\institute{Zhejiang University\\
\and
Hangzhou City University\\
\email{\{xiaozhb, yu.wang, liushunyu\}@zju.edu.cn,  \\
\{huiqiong\_wang, brooksong\}@zju.edu.cn,\\doujiang\_zheng@163.com}
}
% \author{Anonymous authors}
\maketitle              % typeset the header of the contribution
\begin{abstract}
\zty{
The burdensome training costs on large-scale graphs have aroused significant interest in graph condensation, which involves tuning Graph Neural Networks (GNNs) on a small condensed graph for use on the large-scale original graph.
Existing methods primarily focus on aligning key metrics between the condensed and original graphs, such as gradients, output distribution and trajectories of GNNs, \lsy{yielding} satisfactory performance on downstream tasks.
However, these complex metrics necessitate intricate external parameters and can potentially disrupt the optimization process of the condensation graph, making the condensation process highly demanding and unstable.
\xzb{
Motivated by the recent success of simplified models across various domains, we propose a simplified approach to metric alignment in graph condensation, aiming to reduce unnecessary complexity inherited from intricate metrics.
We introduce the Simple Graph Condensation (SimGC) framework, which aligns the condensed graph with the original graph from the input layer to the prediction layer, guided by a pre-trained Simple Graph Convolution (SGC) model on the original graph. Importantly, SimGC eliminates external parameters and exclusively retains the target condensed graph during the condensation process.}
This straightforward yet effective strategy achieves a significant speedup of up to 10 times compared to existing graph condensation methods while performing on par with state-of-the-art baselines.
Comprehensive experiments conducted on seven benchmark datasets demonstrate the effectiveness of SimGC in prediction accuracy, condensation time, and generalization capability.
Our code is available at \url{https://github.com/BangHonor/SimGC}.
}
\keywords{Large-scale Graphs \and Graph Neural Networks \and Graph Condensation.}
\end{abstract}
\section{Introduction}

\begin{figure}[t]
\begin{minipage}[t]{0.65\linewidth}
\centering
{\includegraphics[width=1.0\linewidth]{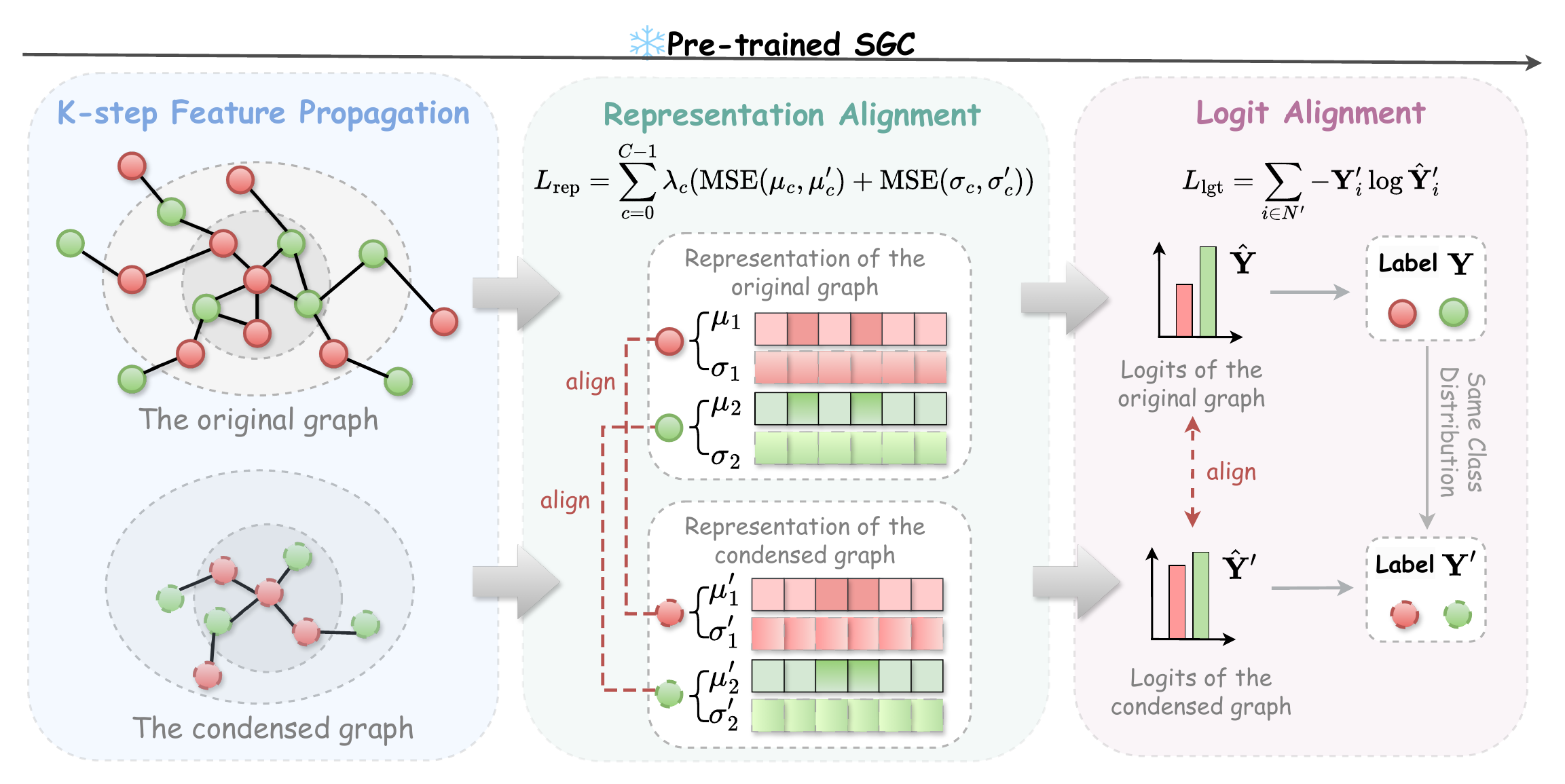}}
(a)
\end{minipage}%
\begin{minipage}[t]{0.35\linewidth}
\centering
{\includegraphics[width=1\linewidth]{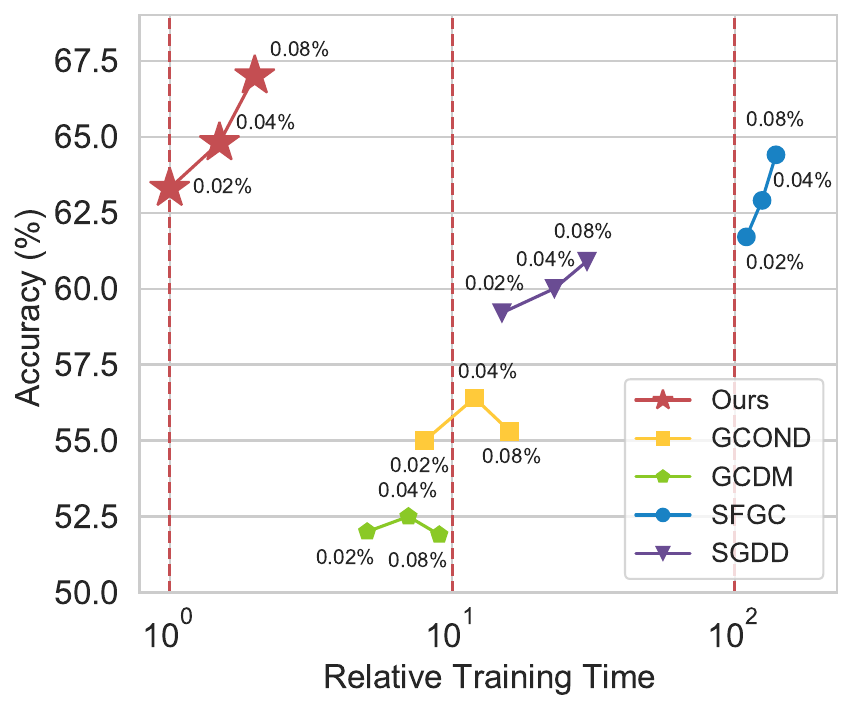}}
(b)
\end{minipage}
\caption{\wy{(a) \xzb{Our proposed SimGC framework operates in two stages. First, an SGC model is pre-trained on the original graph. Subsequently, we align the condensed graph with the original graph through representation alignment in the aggregation layers and logit alignment in the output layer, together with a feature smoothness
regularizer.}
(b) Overall performance of all methods on the largest Ogbn-products dataset, where ``0.02\%", ``0.04\%", ``0.08\%" refer to the reduction rates.}}
\label{fig:introduction}
\end{figure}

The proliferation of mobile devices worldwide has led to the emergence of large-scale graphs, connecting various entities through semantic relationships~\cite{reiser2022graph,wu2020comprehensive,zhou2020graph}. 
In the era of Deep Learning, Graph Neural Networks (GNNs) have revolutionized the learning paradigm for non-Euclidean graph data, providing dense representations for entities in social networks~\cite{brody2021attentive,wu2020rumor}, recommender systems~\cite{fan2019graph,ying2018graph}, fraud detection~\cite{cheng2020graph,zhang2018link}, power systems~\cite{liu2024MAM,liu2024PAC}, and molecule representations~\cite{xu2018powerful,ying2021transformers}.
However, despite the significant achievements in improving prediction accuracy, GNNs encounter computational challenges when dealing with large-scale graphs in industrial settings, hindering their practical applications. Consequently, researchers have proposed promising directions to address these computational costs, including Sampling methods~\cite{hamilton2017inductive,chiang2019cluster}, Distributed Graph Learning~\cite{wan2021pipegcn,peng2022sancus}, and Graph Condensation~\cite{jin2021graph,zheng2023structure}.

\zty{
Graph condensation aims to produce a compact yet informative representative of a large-scale graph, which can be used to train GNNs and applied to the original graph without loss of information.
The pioneering works such as GCOND~\cite{jin2021graph} and DosCond~\cite{jin2022condensing} align the GNN gradients between the condensed graph and the original graph, achieving superior performance to graph sparsification~\cite{peleg1989graph,spielman2011spectral} and graph coarsening~\cite{loukas2018spectrally,loukas2019graph} methods.
Considering additional informative metrics, SGDD~\cite{yang2023does} emphasizes the structural aspects of the original graph, while SFGC~\cite{zheng2023structure} focuses on the training trajectories of GNNs within the original graph, resulting in improved performance at the expense of increased condensation costs.
Apart from these approaches, GCDM~\cite{liu2022graph} optimizes the condensed graph by incorporating a trainable GNN and aligning the output logits of both graphs, trading off the condensation speed for prediction accuracy.
By treating the condensed graph as optimizable parameters, existing methods have explored various metrics (such GNN gradients, output logits and trajectories mentioned above) to effectively align the condensed graph with the original graph.
}

\xzb{However, the utilization of these complex metrics in existing graph condensation incurs high costs due to the inclusion of numerous external parameters (such as trainable GNN parameters), rendering them impractical for large-scale graphs.}
First, these metrics are highly sensitive to high-order computations, which undermines the stability of the condensation process.
Second, the computation complexity of metrics directly influences the condensation costs, resulting in significantly longer training times.
Consequently, condensed graphs generated by these metrics may not yield satisfactory performance in downstream applications.
In recent years, the success of simplifying burdensome models, such as employing a simple MLP or linear model, has been demonstrated in GNNs~\cite{wu2019simplifying,he2020lightgcn,wu2023sgformer}, Computer Vision~\cite{tolstikhin2021mlp}, and Time Series Analysis~\cite{Zeng2022linear}, achieving comparable performance to state-of-the-art (SOTA) models with a minimal number of parameters.
\xzb{
Motivated by this observation, we propose adopting a Simple Graph Convolution network (SGC)~\cite{wu2019simplifying} that aligns with the inductive bias of GNNs for metric alignment, rather than relying on intricate parameters~\cite{wu2019simplifying,he2020lightgcn,wu2023sgformer}.}

\zty{
Therefore, this paper presents a simplified framework, named Simple Graph Condensation (SimGC), that eliminates external parameters and exclusively maintains the condensed graph as parameters throughout the condensation process.
Initially, an SGC is pre-trained on the original graph to capture the rich semantics among features, structures, and labels, which incurs minimal computational costs due to its simplification scheme.
As shown in Fig.~\ref{fig:introduction}(a), we employ a layer-wise alignment strategy, guided by the pre-trained SGC, from the input layer to the output layer.
This ensures that the representation layers, including the inputs of the condensed graph, are aligned with the corresponding layers of the original graph. Additionally, the output layer of the condensed graph is aligned with the output layer of the original graph. The alignment process plays an integral role in creating a semantic concordance between node features, adjacency matrices and node labels of the condensed and original graphs. As a result, it enables the condensed graph to maintain similar capabilities to effectively train GNNs like the original graph.
Furthermore, drawing inspiration from the concept of adjacent similarity in graphs~\cite{wu2019adversarial}, we introduce a smoothness regularizer to enhance the interdependent learning of features and structures within the condensed graph.
With its simplified parameters, Fig.~\ref{fig:introduction}(b) shows that SimGC achieves a significant acceleration in condensation speed, up to 10 times faster, while maintaining comparable or even higher accuracy to SOTA counterparts.
Comprehensive experiments conducted on seven benchmark datasets further validate the effectiveness of SimGC in terms of prediction accuracy, condensation time, and generalization capability.
The contributions of this work are summarized as follows:
}
\begin{itemize}
[nosep,leftmargin=1em,labelwidth=*,align=left]
\item[$\bullet$] \textbf{Methodology}: 
\zty{
We propose a simplified framework called Simple Graph Condensation (SimGC) to streamline the graph condensation process by removing external parameters associated with existing methods.
}
% In this paper, we propose a simple yet effective graph condensation framework termed as Simple Graph Condensation (SimGC), which eliminates external parameters and exclusively retains the target condensed graph during the condensation process, greatly reducing the computation cost and optimization difficulty.
\item[$\bullet$] \textbf{Efficiency}:
\zty{
SimGC demonstrates a consistent speed improvement across different graphs when compared to existing methods, achieving a significant speedup of up to 10 times compared to existing graph condensation methods.
}
% The condensation time cost of SimGC is much less than other baselines. In particular, the speed of condensation with SimGC demonstrates a minimum of
% two times acceleration in the process. Notably, on the Ogbn-products dataset, the total time is reduced by an impressive factor of five.
\item[$\bullet$] \textbf{Accuracy}: 
\zty{
Extensive experiments have convincingly proved that SimGC performs on par with SOTA methods across all datasets.
Notably, SimGC exhibits superiority over existing SOTA methods on the Ogbn-products, Reddit and Reddit2 datasets at all reduction rates.
}
% Baseline comparison results demonstrate that SimGC outperforms other condensing methods across majorities of datasets with extremely low cost compared to other baselines. In particular, our method outperforms existing SOTA techniques by over 1\% and 2\% on Reddit and Reddit2 datasets, respectively.
% \item \textbf{Application}: The condensed graph can be applied to distill a student GNN from the teacher GNN and obtain a high-performance student GNN, at the same time remarkably speeding up the KD process. Additionally, the Neural Architecture Search (NAS) result of the condensed graph and the original graph is very close, while the search speed of the condensed graph is way faster than the original graph, allowing us to use the condensed graph to speed up the NAS process.
\end{itemize}

\section{RELATED WORK}
\noindent\textbf{Dataset Condensation.}
Dataset distillation and dataset condensation~\cite{wang2018dataset} aims to synthesize small datasets from the original data while maintaining similar performance levels of models trained on them. Zhao et al.~\cite{zhao2020dataset} propose a data condensation method that matches gradients with respect to model parameters between the condensed and original datasets. Zhao et al. \cite{zhao2023dataset} also introduce a technique to reduce computational costs by matching the distribution of the two datasets and condensing the original dataset. Additionally, Nguyen et al~\cite{nguyen2020dataset}. condense datasets by utilizing Kernel Inducing Points (KIP) and approximating neural networks with kernel ridge regression (KRR). Wang et al.~\cite{wang2022cafe} propose CAFE, which enforces consistency in the statistics of features between synthetic and real samples extracted by each network layer, except the final one. 
Liu et al.~\cite{liu2023slimmable} propose a novel training objective for slimmable dataset condensation, which extracts a smaller synthetic dataset given only previous condensation results. 
These approaches demonstrate various techniques for condensing datasets while maintaining substantial performance. By leveraging the condensed dataset, training processes can be accelerated, and storage efficiency can be improved, leading to significant enhancements in tasks such as NAS \cite{such2020generative}, continual learning \cite{deng2022remember}, and differential privacy \cite{dong2022privacy}. However, Applying dataset condensation methods to graphs is tough due to their dual node and edge nature. Traditional methods focus mainly on features, but graphs also involve adjacency matrices. The interaction between these matrices and node features makes graph condensation complex and computation-heavy.

\noindent\textbf{Graph Condensation.}
% \subsection{Graph Condensation}
% Traditional methods for graph size reduction include graph sparsification~\cite{spielman2011spectral, peleg1989graph} and graph coarsening~\cite{loukas2019graph, loukas2018spectrally}. However, these methods often sacrifice performance by directly discarding a significant portion of the node and edges. 
Graph condensation, similar to dataset condensation, aims to synthesize a smaller graph that effectively represents the original graph for training GNNs. GCOND~\cite{jin2021graph} achieves graph condensation by minimizing the gradient-matching loss between the gradients of training losses with respect to the GNN parameters of both the original and condensed graphs. DosCond~\cite{jin2022condensing} and SGDD~\cite{yang2023does} are also based on gradient-matching, the former simplifies the gradient-matching process while the latter proposes to broadcast the original structure information to the condensed graph to maintain similar structure information. 
SFGC~\cite{zheng2023structure} synthesizes a condensed graph by matching the training trajectories of the original graph.
GCDM~\cite{liu2022graph} takes a distribution-matching approach by treating the original graph as a distribution of reception fields of specific nodes.  
% GC-SNTK~\cite{wang2023fast} propose reforming the graph condensation problem as a Kernel Ridge Regression (KRR) task. 
While these techniques demonstrate superiority over conventional methods in certain situations, they primarily focus on aligning key metrics between the condensed and original graphs, such as gradients, output distribution and trajectories of GNNs. The convoluted metrics involved necessitate complex computations and can potentially disrupt the optimization process of the condensation graph, making the condensation process highly demanding and unstable.

\section{Problem Definition}
\zty{
The goal of graph condensation is generating a small condensed graph when given a large-scale graph, which can be used for burdensome parameter tuning, GNNs ensemble learning, neural architecture search, etc.
Formally, let $\mathcal{T}=(\mathbf{A}, \mathbf{X}, \mathbf{Y})$ be the original graph dataset and $N$ denote the number of nodes, where $\mathbf{X}\in{\mathbb{R}^{N\times d}}$ is the $d$-dimensional node feature matrix, $\mathbf{A}\in \mathbb{R}^{N\times N}$ is the adjacency matrix, $\mathbf{Y}\in\{0,\ldots,C-1\}^N$ represents the node labels over $C$ classes.
The condensed graph is denoted by $\mathcal{S}=({\mathbf{A}'}, {\mathbf{X}'},{\mathbf{Y}'})$ sharing the same graph signature with the original graph $\mathcal{T}$, where ${ \mathbf{A}'}\in\mathbb{R}^{N'\times N'}$, ${ \mathbf{X}'}\in\mathbb{R}^{N'\times d}$, ${ \mathbf{Y}'}\in\{0,\ldots, C-1\}^{N'}$ and $N'\ll{N}$.
In this way, arbitrary GNNs trained on the condensed graph $\mathcal{S}$ can be seamlessly applicable to tasks of the original graph $\mathcal{T}$, while offering negligible training costs.
The optimization objective of the graph condensation process can be formulated as follows:
}
\begin{gather}
{\min_{\mathcal{S}}} \; \mathcal{L}\left(\text{GNN}_{{\theta}_{\mathcal{S}}}(\mathbf{ A},\mathbf{ X}), \mathbf{ Y}\right) \nonumber \quad \\ 
\text { s.t } \quad {\theta}_{\mathcal{S}}=\underset{{\theta}}{\arg \min } \; \mathcal{L}(\text{GNN}_{{\theta}}(\mathbf{A}',\mathbf{X}'), \mathbf{Y}'),
\label{GraphCondensation}
\end{gather}
\xzb{where $\text{GNN}_{{\theta}_{\mathcal{S}}}$ represents the GNN model trained on the condensed graph $\mathcal{S}$, and $\mathcal{L}$ denotes the training loss function for GNNs, for instance, the cross-entropy loss.}
Directly optimizing the objective function in this bi-level optimization problem is challenging. The formula involves the second-order derivatives with respect to the parameters of GNNs, which is computationally expensive and can lead to numerical instability in deep learning models. Furthermore, the various architectures of GNNs make it even more challenging to compute.
Therefore, existing methods concentrate on approximating the objective by optimizing the key metrics (such as GNN gradients, output logits and trajectories) between the condensed graph and the original graph.

\section{METHOD}

% \begin{figure*}[ht]
% \centering % 图片居中对齐
% \includegraphics[width=0.9\linewidth]{framework_5.pdf}
% \caption{\zty{Illustration of the proposed SimGC. Firstly, we capture the semantics of the original graph with an extremely lightweight pre-trained SGC~\cite{wu2019simplifying}. Secondly, we align $K$-order representations and output logits of the condensed graph with the original graph, following the hierarchical aggregation principles of GNNs. Finally, we solely optimize the parameters of the condensed graph together with a smoothness regularizer, benefiting the neighborhood homophily of the condensed graph.}}
% \label{fig:framework}
% \end{figure*}

\xzb{To address the issues caused by complex metrics of existing methods, we propose adopting an SGC network for metric alignment, which aligns with the inductive bias of GNNs while eliminating the need for intricate external parameters.} Our methodology focuses on aligning important metrics at each layer of the SGC model pre-trained on the original graph, through the proposed representation and logit alignment. The alignment process ensures a significant semantic alignment between node features, adjacency matrices and node labels of the original and condensed graphs, indicating that they possess a similar capability to train GNNs. Additionally, we employ a feature smoothness regularizer to promote neighborhood homophily within the condensed graph, further enhancing its performance.
\zty{Fig.~\ref{fig:introduction}(a) illustrates the core idea of SimGC.}

\subsection{The Condensed Graph}
One essential challenge in the graph condensation problem is how to model the condensed graph and resolve dependency among nodes. We follow the work of GCOND~\cite{jin2021graph} to treat $\mathbf{X}'$ as free parameters, at the same time keeping the class distribution of $\mathbf{Y}'$ the same as the one of $\mathbf{Y}$. 
We introduce ${r}$ as the reduction rate of the condensed graph. This means that if the node number of $c$ class in the original graph is $N_{c}$, then the node number of the corresponding class in the condensed graph will be ${N_{c}{r}}$.
% \begin{equation}
% V'_{c}=\lceil {V_{c}{r}} \rceil, \ \text{for} \ c = 0,...,C-1.
% \label{computeY}
% \end{equation}
As for $\mathbf{A}'$, Jin~\cite{jin2021graph} proves that there is a strong connection between the features and the adjacency matrix. Therefore, we parameterize the adjacency matrix $\mathbf{A}'$ as a function of $\mathbf{X}'$, represented as $G_{\phi}(\mathbf{X}')$ with:
\begin{equation}
a'_{ij}=\operatorname{Sigmoid}\left(\frac {1}{2} \cdot \left( g_{\phi}([\mathbf{X}'_{i};\mathbf{X}'_{j}])+g_{\phi}([\mathbf{X}'_{j};\mathbf{X}'_{i}])\right)\right),
\label{computea}
\end{equation}
\begin{equation}
\mathbf{A}'_{ij}=\left\{
\begin{aligned}
a'_{ij}\ , \ \text{if} \ a'_{ij}\ {\geq}\ {\delta}  \\
0\ ,\ \text{otherwise},
\end{aligned}
\right.
\label{computeA}
\end{equation} where $g_{\phi}$ is an MLP for nonlinear transformation, $[\mathbf{X}'_{i};\mathbf{X}'_{j}]$ denotes the concatenation of the $i$-th and $j$-th nodes features, $\mathbf{A}'_{ij}$ denotes the edge weight of the $i$-th and $j$-th nodes, ${\delta}$ is the hyperparameter that governs sparsity and computation reduction. Consequently, our primary objective involves optimizing $\mathbf{X}'$ and the MLP parameters ${\phi}$ concerning the objective function.

\subsection{The Pre-trained SGC} 
Traditional GNNs usually adhere to a Message Passing paradigm, where non-linear representations are achieved through continuous aggregation of information from neighbors. In such a context, as long as the condensed graph and the original graph remain consistent in layerwise semantics from the input to the output layer, GNNs trained on the condensed graph should be effortlessly applied to the original graph.
Simple Graph Convolution network (SGC)~\cite{wu2019simplifying} is a minimalist style of Message Passing Neural Network (MPNN), characterized by its absence of non-linear transformation parameters during aggregation (feature propagation) steps. This unique feature allows SGC to obtain each layer's semantic at an extremely low cost and with high speed, at the same time maintaining a predictive capability comparable to traditional GNNs. Therefore, a pre-trained SGC on the original graph can thoroughly guide the condensed graph in learning the layerwise semantic of the original graph, making it particularly suitable for graph condensation. The process of pre-training SGC is as follows:
\begin{itemize}
\item[(1)] Obtain normalized adjacency matrix $\hat{\mathbf{A}}$ after adding self-loops of the original graph.
\item[(2)] Acquire $\hat{\mathbf{A}}^{K}\mathbf{X}$ by performing $K$-step feature propagation on $\mathbf{X}$ with $\hat{\mathbf{A}}$.
\item[(3)] Obtain the SGC output $\hat{\mathbf{A}}^K \mathbf{X} \mathbf{\Theta}$ by transforming $\hat{\mathbf{A}}^{K}\mathbf{X}$ with a prediction layer $\mathbf{\Theta}$.
\item[(4)] Optimize the cross-entropy loss $L_{\mathbf{\Theta}} =\mathcal{L}(\operatorname{Softmax}(\hat{\mathbf{A}}^K \mathbf{X}\mathbf{\Theta}), \mathbf{Y})$.
% \begin{equation}
% {L_{\mathbf{\Theta}}}=\mathcal{L}(\operatorname{Softmax}(\hat{\mathbf{A}}^K \mathbf{X}\mathbf{\Theta}), \mathbf{Y}),
% \label{CrossEntropy}
% \end{equation}
% where $\mathcal{L}$ is a loss function such as cross-entropy. 
\end{itemize}
% Firstly, we obtain normalized adjacency matrix $\hat{A}$ with added self-loops of the original graph.
% \begin{equation}
% \hat{A}=\tilde{{D}}^{-\frac{1}{2}} \tilde{{A}} \tilde{{D}}^{-\frac{1}{2}},
% \label{normalized}
% \end{equation}
% where $\tilde{{A}}={A}+{I}$ and $\tilde{{D}}$ is the degree matrix of $\tilde{{A}}$.
% Then we can aggregate the feature $\mathbf{X}$ for $K$ times by repeating $K$ multiplication with $\hat{A}$. We further transform the $\hat{A}^{K}X$ by applying the linear transformation layer $f_{\mathbf{\Theta}}$ and obtain the SGC output $\hat{A}^K X {f_{\theta}}$. 
% The resulting classifier becomes:
% \begin{equation}
% \hat{{Y}}=\operatorname{Softmax}\left(\hat{A}^K {X}{f_{\theta}}\right).
% \end{equation}
% We can pre-train such an SGC by:

\subsection{Representation Alignment} 
The condensed graph $\mathcal{S}$ consists of three variables $(\mathbf{X}', \mathbf{A}', \mathbf{Y}')$, and we already parameterize $\mathbf{A}'$ as $G_{\phi}(\mathbf{X}')$ as outlined in Eq.~\ref{computea} and Eq.~\ref{computeA}. During the representation alignment, we aim to align node features $\mathbf{X}'$ and adjacency matrix $\mathbf{A}'$ with the corresponding variables in the original graph. 
Due to the $K$-step aggregation (feature propagation) property of GNNs, $\mathbf{X}'$ and $\mathbf{A}'$ can be effectively fused using $K$-step representations, enabling simultaneous condensation of both feature and topology by aligning a set of homogeneous representations.

\zty{As illustrated in Fig.~\ref{fig:introduction}(a),} we adopt the $K$-step aggregation rule of traditional GNNs and utilize the alignment of each SGC aggregation step to bridge the gap between the two graphs. Notably, the representation at the 0-th step corresponds to the node feature $\mathbf{X}'$, while the $K$-th step representation is the last layer representation before the prediction layer. For the 0-th step node representation, our goal is to maintain the node representation distribution of the original graph to ensure a similar potential for training GNNs. This is crucial since node distribution governs the essential classification performance of GNNs. For the 1-st to $K$-th step node representations, we need to ensure that the node representation distribution in the corresponding layer of both graphs remains similar, thus ensuring both graphs possess a similar capacity for training GNNs after aggregating with adjacency matrices.
% It is worth noting that the features of $\mathbf{X}'$ are $N'$-dimensional, while the features of $\mathbf{X}$ are $N$-dimensional. Computing the optimal transport distance between both graphs' $K$-step representations would involve computationally expensive operations. Existing methods often reduce the dimensions of GNN parameters or distribution dimensions to enable tractable distance calculation and optimize the small graph. 
We discover that the simple mean and standard deviation statistics of each class can effectively capture the node representation distribution in each layer. Based on this observation, we choose to align the mean and standard deviation of node representations for each class in each layer, allowing us to align the node features and adjacency matrices of both graphs at the same time.
The whole representation alignment process is as follows:
% \begin{enumerate}[leftmargin=0.5cm]
\begin{itemize}
[nosep,leftmargin=1em,labelwidth=*,align=left]
\item[(1)] Perform $K$ times feature propagation on both graphs using the aggregation layers of the pre-trained SGC: $\mathbf{e}_{k}=\hat{\mathbf{A}}^{k}\mathbf{X}$, $\mathbf{e}'_{k}=\hat{\mathbf{A}}'^{k}\mathbf{X}'$.
% \begin{equation}
% e_{i}=\hat{A}^{i}X,
% \end{equation}
% \begin{equation}
% e'_{i}=\hat{A'}^{i}X'.
% \end{equation}
Let $\mathbf{e}_{k}$ and $\mathbf{e}'_{k}$ denote the node representations of the original graph and the condensed graph after the $k$-th aggregation steps. $\hat{\mathbf{A}}'$ is the normalized adjacency matrix of the condensed graph.
\item[(2)] Concat the representations from 0-layer to $K$-layer and form the concatenated $K$-step representations: $\mathbf{Z}=[\mathbf{X}; \mathbf{e}_{1};...; \mathbf{e}_{K}]$, $\mathbf{Z}'=[\mathbf{X}'; \mathbf{e}'_{1};...; \mathbf{e}'_{K}]$. 
The concatenated $K$-step representations of $c$ class in both graphs can be represented as $\mathbf{Z}_c$ and $\mathbf{Z}'_c$. 
\item[(3)] Compute the representation alignment loss as:
\begin{equation}
L_{\text{rep}}=\sum_{c=0}^{C-1}\lambda_c(\text{MSE}(\mathbf{\mu}_c, \mathbf{\mu}'_c)+\text{MSE}(\mathbf{\sigma}_c, \mathbf{\sigma}'_c)),
\label{alignment}
\end{equation}
where $\mathbf{\mu}_c$ and $\mathbf{\mu}'_c$ denote the means of $\mathbf{Z}_c$ and $\mathbf{Z}'_c$. $\mathbf{\sigma}_c$ and $\mathbf{\sigma}'_c$ denote the standard deviations of $\mathbf{Z}_c$ and $\mathbf{Z}'_c$ respectively. The proportion $\lambda_c$ of each class is computed by dividing the node number of $c$-th class by that of the size of the largest class. And ``MSE" is the Mean Squared Error loss function.
% \end{enumerate}
\end{itemize}
% By using the representation alignment, we can ensure that both graphs exhibit high similarity in feature distribution and topology structure, making it easier for the condensed graph to train GNNs that are comparable to the original graph.

\subsection{Logit Alignment}
% The representation alignment module initially prioritizes the mutual influence between $\mathbf{X}'$ and $\mathbf{A}'$. 
% In this section, we further explore the mutual influence of $\mathbf{X}'$, $\mathbf{A}'$, and $\mathbf{Y}'$ and aim to align them with the corresponding variables in the original graph using the output logit of the SGC.
% We observe that the output logit is a composite result of node features and topological structure. Given that: (1) The output logit generated by $\mathbf{X}$ and $\mathbf{A}$ exhibits a strong alignment with the original labels $Y$ when using the pre-trained SGC model. (2) $\mathbf{Y}'$ is sampled from $\mathbf{Y}$, and they share the same class distribution, implying that there is alignment between $\mathbf{Y}$ and $\mathbf{Y}'$.
% We propose that the output logits generated by $\mathbf{X}'$ and $\mathbf{A'}$ should align with the output logits produced by $\mathbf{X}$ and $\mathbf{A}$ by maintaining consistency with the corresponding labels under the same pre-trained SGC model. This alignment process facilitates the effective transfer of knowledge acquired from training on the original graph to the condensed graph, enabling $\mathbf{X}'$ and $\mathbf{A}'$ to inherit the GNN training abilities of $\mathbf{X}$ and $\mathbf{A}$, which is in line with the principles of knowledge distillation \cite{liu2023graph}. The logit alignment loss is computed as follows:
% \begin{equation}
% \mathcal{L}_\text{lgt} = \mathcal{L}(\operatorname{Softmax}(\hat{\mathbf{A}'}^K \mathbf{X}'\mathbf{\Theta}), Y').
% \label{logit}
% \end{equation}
The condensed graph $\mathcal{S}$ consists of three variables $(\mathbf{X}', \mathbf{A}', \mathbf{Y}')$, the above representation alignment module focuses on aligning $(\mathbf{X}', \mathbf{A}')$ with the original graph.
In this section, we propose to align the output logits $\hat{\mathbf{Y}}$ of the original graph and the output logits $\hat{\mathbf{Y}}'$ of the condensed graph at the output layer. We accomplish this by maintaining the consistency between $\hat{\mathbf{Y}}'$ and the corresponding labels $\mathbf{Y}'$ under the same pre-trained SGC, mirroring the relationship between  $\hat{\mathbf{Y}}$ and $\mathbf{Y}$. Here, $\hat{\mathbf{Y}}'=\operatorname{Softmax}(\hat{\mathbf{A}}'^K \mathbf{X}'\mathbf{\Theta})$, and $\mathbf{\Theta}$ is already pre-trained in the SGC. Moreover, $\mathbf{Y}'$ is sampled in proportion to the original graph $\mathbf{Y}$, indicating they have the same class distribution.
The logit alignment loss is computed as follows:
\begin{equation}
{L_\text{lgt}}= \sum_{i \in N'} - \mathbf{Y}'_i \log \hat{\mathbf{Y}}'_i.
\label{eq:logit}
\end{equation}
Firstly, this kind of logit alignment connects the node features and adjacency matrix with the sampled node labels, facilitating the condensation consistency of three variables.
Secondly, it also simulates the knowledge learning~\cite{liu2023graph} of the original graph by distilling the knowledge $\mathbf{\Theta}$ of the original graph into the condensed graph, which is achieved by mimicking the behaviors of prediction logits $\hat{\mathbf{Y}}$ of the original graph.
Therefore, our proposed logit alignment works at the output layer to align both graphs, enabling the condensed graph to inherit the GNN training abilities of the original graph.

\subsection{Optimization}
In summary, with the above representation alignment and logit alignment, SimGC can learn the node feature, topology structure, and label distribution of the original graph fully under the guidance of the pre-trained SGC. By eliminating the need for external training parameters, SimGC ensures a highly efficient condensation process. Moreover, it significantly reduces the optimization complexity, resulting in superior performance on the condensed graph for various downstream tasks.

However, it is important to note that while we have utilized $\mathbf{X}'$ to construct $\mathbf{A}'$, $\mathbf{A}'$ could also potentially impact $\mathbf{X}'$ in return. In numerous real-world graph data such as social networks and citation networks, connected nodes tend to possess similar features~\cite{wu2019adversarial}.
Therefore, it is crucial to ensure that the condensed graph maintains this property. We propose using a kernel function to map the features to a higher dimension and calculate their similarity. As a result, we include a loss term $L_\text{smt}$ to consider the feature smoothness of the condensed graph:
\begin{equation}
L_\text{smt}=\sum_{i,j=1}^{N'}{\mathbf{A}'_{ij}{\left \langle \varphi(\mathbf{X}'_{i}), \varphi(\mathbf{X}'_{j}) \right \rangle}},
\label{smoothness}
\end{equation} where $\varphi(.)$ represents a kernel function, and the mapped vectors' similarity is computed as their inner product. Exploiting kernel trick enables us to calculate feature similarity more accurately without requiring explicit knowledge of $\varphi$. 
Here we utilize the radial basis function (RBF) as our kernel function: $\langle \varphi(\mathbf{X}'_{i}), \varphi(\mathbf{X}'_{j}) \rangle =e^{-{\frac{1}{2\sigma^2}\Vert \mathbf{X}'_{i}-\mathbf{X}'_{j}} \Vert^2}$.
% The radial basis function (RBF), linear function (Linear), and polynomial kernel function (Poly) are among the most commonly used kernel functions:
% \begin{equation}
% {\left \langle \varphi(X'_{i}), \varphi(X'_{j}) \right \rangle}=\left\{
% \begin{aligned}
% &e^{-{\frac{1}{2\sigma^2}\Vert X'_{i}-X'_{j}} \Vert^2} \quad \ \text{\bf RBF}\\
% &{{X'_{i}}^{T}}{X'_{j}} \qquad \qquad \ \ \ \text{\bf Linear}\\
% &({{X'_{i}}^{T}}{X'_{j}}+c)^d  \qquad \text{\bf Poly},
% \end{aligned}
% \right.
% \label{kernel}
% \end{equation}
% \begin{equation}
% \langle \varphi(X'_{i}), \varphi(X'_{j}) \rangle =e^{-{\frac{1}{2\sigma^2}\Vert X'_{i}-X'_{j}} \Vert^2}
% \label{kernel}
% \end{equation}

Finally, the overall condensation loss is as follows: 
\begin{equation}
{\min_\mathcal{S}L}= \alpha L_\text{rep} + \beta L_\text{lgt} + \gamma L_\text{smt},
\label{objective}
\end{equation}
where $L_\text{rep}$ is the representation alignment loss, $L_\text{logit}$ is the logit alignment loss and $L_\text{smt}$ is the feature smoothness loss. $\alpha$, $\beta$, $\gamma$ are hyperparameters that control the weights of the above three losses respectively.
The overall framework is provided in Algo.~\ref{alg:simgc}.
\begin{algorithm}[t]
\renewcommand{\arraystretch}{0.8}
\raggedright
\SetAlgoVlined
\scriptsize
\textbf{Input:} Original graph $\mathcal{T}=(\mathbf{A}, \mathbf{X}, \mathbf{Y})$, an SGC model with a prediction layer $\mathbf{\Theta}$ pre-trained on $\mathcal{T}$\\
Obtain $K$-step representations $\mathbf{Z}$ of $\mathcal{T}$ \\
% \hfill$\triangleright$ Eq.~(\ref{eq:aggregate})
Randomly initialize the condensed graph $\mathcal{S}=(\mathbf{X}', G_{\phi}(\mathbf{X}'), \mathbf{Y}')$\\
\For{$t = 0,\ldots,T-1$}
{   
    $\mathbf{A}'=G_{\phi}(\mathbf{X}')$\\
    Obtain $K$-step representations $\mathbf{Z}'$ of $\mathcal{S}$ \\
    % \hfill$\triangleright$ Eq.~(\ref{eq:aggregate})
    $L_\text{rep} = 0$\\
    \For{$c = 0,\ldots,C-1$}
    {   
        Obtain the mean $\mathbf{\mu}_c$ and standard deviation $\mathbf{\sigma}_c$ of $\mathbf{Z}_c$ \\
        Obtain the mean $\mathbf{\mu}'_c$ and standard deviation $\mathbf{\sigma}'_c$ of $\mathbf{Z}'_c$ \\
        Obtain the proportion $\lambda_c$ of the $c$-th class \\
        $L_\text{rep} += \lambda_c (\text{MSE}(\mathbf{\mu}_c, \mathbf{\mu}_c') + \text{MSE}(\mathbf{\sigma}_c, \mathbf{\sigma}'_c))$ \\
    }
    $\hat{\mathbf{Y}}'=\operatorname{Softmax}(\hat{\mathbf{A}}'^K \mathbf{X}'\mathbf{\Theta})$\\
    ${L_\text{lgt}}= \sum_{i \in N'} - \mathbf{Y}'_i \log \hat{\mathbf{Y}}'_i$\\
    % ${L_\text{lgt}}=\mathcal{L}(\operatorname{Softmax}(\hat{\mathbf{A}}'^K \mathbf{X}'\mathbf{\Theta}), \mathbf{Y}')$\\
    $L_\text{smt}=\sum_{i,j=1}^{N'}{\mathbf{A}'_{ij}{\left \langle \varphi(\mathbf{X}'_{i}), \varphi(\mathbf{X}'_{j}) \right \rangle}}$\\
    $L = \alpha L_\text{rep}+ \beta L_\text{lgt}+ \gamma L_\text{smt}$\\
    \If{$t \% (\tau_1+\tau_2) < \tau_1$}{
        Update $ \mathbf{X}'=\mathbf{X}'-\eta_{1}\bigtriangledown_{\mathbf{X}'}L$
    }
    \Else{
        Update $\phi=\phi- \eta_{2}\bigtriangledown_{\phi}L$
    }
}
$\mathbf{A}'=G_{\phi}(\mathbf{X}')$\\
\textbf{Output:} $\mathcal{S}=(\mathbf{A}', \mathbf{X}', \mathbf{Y}')$
\caption{The proposed SimGC framework}
\label{alg:simgc}
\end{algorithm}

\section{Experiments}
% In this section, we perform experiments to validate the effectiveness of the proposed framework. We begin by describing the experimental settings and subsequently comparing SimGC to representative baselines in terms of performance, time, and generalizability. Detailed discussions on the results are also provided.

\subsection{Experiment Settings}
\noindent\textbf{Datasets.} We evaluate the condensation performance of our method on seven datasets, four of which are transductive datasets such as Cora and Citeseer~\cite{kipf2016semi}, Ogbn-arxiv and Ogbn-products~\cite{hu2020open} and three of which are inductive datasets like Flickr~\cite{zeng2019graphsaint}, Reddit~\cite{hamilton2017inductive} and Reddit2~\cite{zeng2019graphsaint}. We directly use public splits to split them into the train, validation, and test sets. The detailed statistics are summarized in Table~\ref{tab:data}. In the condensation process, we make full use of the complete original graph and training set labels for transductive datasets, whereas we use only the training set subgraph and corresponding labels for inductive datasets.
\begin{table*}[t]
\scriptsize
\centering
\renewcommand{\arraystretch}{0.8}
\caption{The statistics of seven datasets.}
\begin{tabular}{{l@{\hspace{8pt}}l@{\hspace{8pt}}r@{\hspace{8pt}}r@{\hspace{8pt}}r@{\hspace{8pt}}r@{\hspace{8pt}}r@{\hspace{8pt}}r@{\hspace{8pt}}c}}
\toprule
\textbf{Dataset} & \textbf{Node Classification} & \textbf{\#Nodes} & \textbf{\#Edges} & \textbf{\#Classes}  &
\textbf{\#Training}  &
\textbf{\#Validation}  &
\textbf{\#Test}  \\
\midrule  
Cora &  Transductive & 2,708 & 10,556 & 7 & 140 & 500 & 1,000\\
Citeseer &  Transductive  & 3,327 & 4,732 & 6 & 120 & 500 & 1,000\\
Ogbn-arxiv & Transductive  & 169,343 & 1,166,243 & 40  & 90,941 & 29,799 & 48,603 \\ 
Ogbn-products & Transductive  & 2,449,029 & 61,859,140	 & 47  & 196,615 & 39,323 & 2,213,091 \\ \midrule
Flickr &  Inductive & 89,250 & 899,756 & 7 & 44,625 & 22,312 & 22,313\\
Reddit & Inductive & 232,965 & 114,615,892 & 41  & 153,431 & 23,831 & 55,703\\ 
Reddit2 & Inductive & 232,965 & 23,213,838 & 41  & 153,932 & 23,699 & 55,334\\ %\midrule
\bottomrule
\end{tabular}
\label{tab:data}
\end{table*}

\noindent\textbf{Baselines.} We compare our approach with several baseline methods: (1) Three coreset methods including Random, Herding~\cite{welling2009herding} and K-Center~\cite{sener2017active}; Four graph condensation methods including GCOND~\cite{jin2021graph}, GCDM~\cite{liu2022graph}, SFGC~\cite{zheng2023structure} and SGDD~\cite{yang2023does}. 
% For the coreset methods, we obtain a condensed graph from the original graph by selecting a few core nodes with the three corset methods and inducing a subgraph from these nodes. 
Since the source code for GCDM  is not available, we re-implement our own version of GCDM from scratch. The experimental results demonstrate that our implemented GCDM yields consistent results with those reported in the original paper~\cite{liu2022graph}. To ensure reliability, we employ the published results of~\cite{liu2022graph}. For the results that are not provided in their papers, we conduct experiments ourselves with their source code to obtain the results.

\noindent\textbf{Experimental Pipeline.} The experiment process is divided into three stages: (1) obtaining a condensed graph from the original graph with a reduction rate $r$, (2) training a model GNN$_S$ with the condensed graph, then selecting the best-performed model using the original validation set, and (3) evaluating the model with the original test set. 

\noindent\textbf{Hyperparameter Settings.} The SGC we use comprises two aggregation layers and a linear or nonlinear prediction layer. We assess the performance of the condensed graph using a GCN with a two-layer architecture. Most of the experiments are performed using a single Quadro P6000, except for those on Ogbn-products that use a single NVIDIA A40 due to the substantial GPU memory. 
The learning rates for $\mathbf{X}'$ and the MLP parameters ${\phi}$ are set to 0.005 and 0.001 for Cora and Citeseer, and 0.05 and 0.01 for the other datasets.
The hyperparameter ${\delta}$, which governs sparsity, is set to 0.01 for all the datasets. 
% For the transductive datasets Cora, Citeseer, Ogbn-arxiv, and Ogbn-products, the labeling rates are 5.2\%, 3.6\%, 53\%, and 8\%, respectively. We choose reduction rates $r$ of \{25\%, 50\%, 100\%\}, \{25\%, 50\%, 100\%\}, \{0.1\%, 0.5\%, 1\%\}, \{0.25\%, 0.5\%, 1\%\} of the label rates, with corresponding final reduction rates of \{1.3\%, 2.6\%, 5.2\%\}, \{0.9\%, 1.8\%, 3.6\%\}, \{0.05\%, 0.25\%, 0.5\%\}, and \{0.02\%, 0.04\%, 0.08\%\}.
% For the inductive datasets Reddit and Reddit2, the reduction rates are both set to \{0.05\%, 0.1\%, 0.2\%\}. For Flickr, the reduction rates are set to \{0.1\%, 0.5\%, 1\%\}.
% All parameters in this article remain consistent with those mentioned above unless specifically stated otherwise.
\begin{table*}[t]
\scriptsize
\setlength{\tabcolsep}{1.5pt}
\renewcommand{\arraystretch}{0.8}
\center
\caption{The performance comparison of our proposed SimGC and baselines on \wy{seven} datasets under various reduction rates. Performance is reported \wy{by} test accuracy (\%). ``$\pm$" corresponds to \wy{the} standard deviation of the average evaluation over 5 trials. ``Whole Dataset" refers to training with the whole dataset without graph condensation.}
\label{tab:main}
\begin{tabular}{@{}c c ccccccc c c@{}}
\toprule
 & & \multicolumn{7}{c}{Baselines} & \multicolumn{1}{c}{Proposed} & \\
\cmidrule(l{2pt}r{2pt}){3-9} \cmidrule(l{2pt}r{2pt}){10-10}
\multicolumn{1}{c}{Dataset}        & Ratio ($r$) & \begin{tabular}[c]{@{}c@{}}Random\end{tabular} & \begin{tabular}[c]{@{}c@{}}Herding\end{tabular} & \begin{tabular}[c]{@{}c@{}}K-Center\end{tabular} & \begin{tabular}[c]{@{}c@{}}GCOND\end{tabular} & \begin{tabular}[c]{@{}c@{}}GCDM\end{tabular} & \begin{tabular}[c]{@{}c@{}}SFGC\end{tabular} 
& \begin{tabular}[c]{@{}c@{}}SGDD\end{tabular}
& \begin{tabular}[c]{@{}c@{}}SimGC\end{tabular} 
& \multicolumn{1}{l}{\begin{tabular}[c]{@{}c@{}}Whole\\ Dataset\end{tabular}} \\ \midrule
\multirow{3}{*}{\begin{tabular}[c]{@{}c@{}} \textsf{Cora} \end{tabular}} 
& 1.3\%           
& 63.6$\pm$3.7                                                 
& 67.0$\pm$1.3                                                         
& 64.0$\pm$2.3                                                     
& 79.8$\pm$1.3   
& 69.4$\pm$1.3 
& 80.1$\pm$0.4
& 80.1$\pm$0.7
& \pmb{80.8$\pm$2.3} 
& \multirow{3}{*}{81.2$\pm$0.2}     \\
& 2.6\%            
& 72.8$\pm$1.1                                                       
& 73.4$\pm$1.0                                                         
& 73.2$\pm$1.2                                                     
& 80.1$\pm$0.6  
& 77.2$\pm$0.4 
& 81.7$\pm$0.5
& 80.6$\pm$0.8
& \pmb{80.9$\pm$2.6}
&                           \\
& 5.2\%          
& 76.8$\pm$0.1                                                        
& 76.8$\pm$0.1                                                         
& 76.7$\pm$0.1                                                      
& 79.3$\pm$0.3
& 79.4$\pm$0.1 
& 81.6$\pm$0.8
& 80.4$\pm$1.6
& \pmb{82.1$\pm$1.3}
&                           \\ \midrule
\multirow{3}{*}{\textsf{Citeseer}}   & 0.9\%           
& 54.4$\pm$4.4                                                        
& 57.1$\pm$1.5                                                        
& 52.4$\pm$2.8                                                          
& 70.5$\pm$1.2                                              
& 62.0$\pm$0.1                                               
& 71.4$\pm$0.5   
& 69.5$\pm$0.4
& \pmb{73.8$\pm$2.5} 
& \multirow{3}{*}{71.7$\pm$0.1}     \\
& 1.8\%            
& 64.2$\pm$1.7                                                        
& 66.7$\pm$1.0                                                         & 64.3$\pm$1.0                                          
& 70.6$\pm$0.9                                               
& 69.5$\pm$1.1                                                      
& \pmb{72.4$\pm$0.4}
& 70.2$\pm$0.8
& 72.2$\pm$0.5
&                           \\
& 3.6\%              
& 69.1$\pm$0.1                                                        
& 69.0$\pm$0.1                                                         & 69.1$\pm$0.1                                        
& 69.8$\pm$1.4                                               
& 69.8$\pm$0.2                                                      
& 70.6$\pm$0.7   
& 70.3$\pm$1.7
& \pmb{71.1$\pm$2.8}
&                           
\\ \midrule
\multirow{3}{*}{\begin{tabular}[c]{@{}c@{}} \textsf{Ogbn-arxiv} \end{tabular}} 
& 0.05\%          
& 47.1$\pm$3.9                                                       
& 52.4$\pm$1.8                                                         
& 47.2$\pm$3.0                                                             
& 59.2$\pm$1.1                                                     
& 51.0$\pm$2.9
& \pmb{65.5$\pm$0.7}
& 60.8$\pm$1.3
& 63.6$\pm$0.8
& \multirow{3}{*}{71.4$\pm$0.2}     \\
& 0.25\%          
& 57.3$\pm$1.1                                                        
& 58.6$\pm$1.2                                                         
& 56.8$\pm$0.8                                                        
& 63.2$\pm$0.3 
& 59.6$\pm$0.4
& 66.1$\pm$0.4
& 65.8$\pm$1.2
& \pmb{66.4$\pm$0.3}
&                           \\
& 0.5\%            & 60.0$\pm$0.9                                                    & 60.4$\pm$0.8                                                         
& 60.3$\pm$0.4                                                       
& 64.0$\pm$0.4   
& 62.4$\pm$0.1
& 66.5$\pm$0.7
& 66.3$\pm$0.7
& \pmb{66.8$\pm$0.4}
&                           \\  \midrule
\multirow{3}{*}{\begin{tabular}[c]{@{}c@{}} \textsf{Ogbn-products} \end{tabular}} 
& 0.02\%            
& 53.5$\pm$1.3                                                        
& 55.1$\pm$0.3                                                         
& 48.5$\pm$0.2                                                       
& 55.0$\pm$0.8  
& 53.0$\pm$1.9
& 61.7$\pm$0.5
& 57.2$\pm$2.0
& \pmb{63.3$\pm$1.1}
& \multirow{3}{*}{74.0$\pm$0.1}     \\
& 0.04\%            
& 58.5$\pm$0.7                                             
& 59.1$\pm$0.1                                                         
& 53.3$\pm$0.4                                                       
& 56.4$\pm$1.0  
& 53.5$\pm$1.1
& 62.9$\pm$1.2
& 58.1$\pm$1.9
& \pmb{{64.8$\pm$1.1}}
&                           \\ 
& 0.08\%          
& 63.0$\pm$1.2                                                        
& 53.6$\pm$0.7                                                         
& 62.4$\pm$0.5                                                        
& 55.3$\pm$0.3  
& 52.9$\pm$0.9
& 64.4$\pm$0.4
& 59.3$\pm$1.7
& \pmb{67.0$\pm$0.7} \\  \midrule
\multirow{3}{*}{\textsf{Flickr}}     & 0.1\%          
& 41.8$\pm$2.0    
& 42.5$\pm$1.8                                                         & 42.0$\pm$0.7                                                         & 46.5$\pm$0.4
& 46.8$\pm$0.2                                               
& 46.6$\pm$0.2  
& \pmb{46.9$\pm$0.1}  
& 45.3$\pm$0.7                                               
&
\multirow{3}{*}{47.2$\pm$0.1}     \\ 
& 0.5\%          
& 44.0$\pm$0.4                                                       
& 43.9$\pm$0.9                                                        
& 43.2$\pm$0.1                                                        
& 47.1$\pm$0.1                                                          
& \pmb{47.9$\pm$0.3}                                               
& 47.0$\pm$0.1  
& 47.1$\pm$0.3  
& 45.6$\pm$0.4                                              
&                           \\
& 1\%            
& 44.6$\pm$0.2                                                        
& 44.4$\pm$0.6                                                         & 44.1$\pm$0.4                                                         & 47.1$\pm$0.1                                                         & \pmb{47.5$\pm$0.1}                                               
& 47.1$\pm$0.1    
& 47.1$\pm$0.1   
& 43.8$\pm$1.5                                              
&                           \\ \midrule
\multirow{3}{*}{\begin{tabular}[c]{@{}c@{}} \textsf{Reddit} \end{tabular}}
& 0.05\%          & 46.1$\pm$4.4                                                        & 53.1$\pm$2.5                                                         
& 46.6$\pm$2.3                                                     
& 88.0$\pm$1.8  
& 73.9$\pm$2.0
& 89.7$\pm$0.2
& 90.5$\pm$2.1
& \pmb{{91.1$\pm$1.0}}
& \multirow{3}{*}{93.9$\pm$0.1}             \\ 
& 0.1\%          
& 58.0$\pm$2.2                                                        
& 62.7$\pm$1.0                                                         
& 53.0$\pm$3.3                                                         
& 89.6$\pm$0.7 
& 76.4$\pm$2.8 
& 90.0$\pm$0.3
& 91.8$\pm$1.9
& \pmb{{92.0$\pm$0.3}}                                                     
&   \\
& 0.2\%            & 66.3$\pm$1.9                                                        & 71.0$\pm$1.6                                                         
& 58.5$\pm$2.1                                                     
& 90.1$\pm$0.5  
& 81.9$\pm$1.6
& 90.3$\pm$0.3
& 91.6$\pm$1.8
& \pmb{{92.6$\pm$0.1}}
&                          \\ \midrule

\multirow{3}{*}{\begin{tabular}[c]{@{}c@{}} \textsf{Reddit2} \end{tabular}}
& 0.05\%          
& 48.3$\pm$6.4                                                        
& 46.9$\pm$1.2                                                         
& 43.2$\pm$3.2                                                     
& 79.1$\pm$2.2  
& 73.5$\pm$4.7
& 84.4$\pm$1.7
& 86.7$\pm$0.8
& \pmb{{89.6$\pm$0.6}}
& \multirow{3}{*}{93.5$\pm$0.1}             \\ 
& 0.1\%          
& 57.8$\pm$3.1                                                       
& 62.5$\pm$2.8   
& 51.9$\pm$0.7 
& 82.4$\pm$1.0 
& 75.4$\pm$1.8 
& 88.1$\pm$1.9
& 85.8$\pm$1.1
& \pmb{{90.6$\pm$0.3}}                                                     
&   \\
& 0.2\%            
& 65.5$\pm$2.5                                                        
& 71.4$\pm$1.6                                                         
& 57.4$\pm$1.8                                                     
& 80.6$\pm$0.4  
& 80.8$\pm$3.1
& 88.6$\pm$1.1
& 85.4$\pm$0.6
& \pmb{{91.4$\pm$0.2}}                          
&                           \\ \bottomrule 
\end{tabular}
\end{table*}

\subsection{Prediction Accuracy}
To evaluate the effectiveness of the condensed graph of different methods. We report the test accuracies for each method in Table~\ref{tab:main} and make the following observations:

\noindent$\blacktriangleright $ \textbf{Observation 1.} SimGC achieves the most promising performance across the vast majority of the datasets and reduction rates. It outperforms existing SOTA methods on Ogbn-products, Reddit and Reddit2 at \emph{all} ratios $r$. 
% This can potentially be attributed to the capability of SimGC to significantly ease optimization difficulty, leading to a more informative compressed graph that delivers higher performance.

\noindent$\blacktriangleright$ \textbf{Observation 2.} The condensed graph produced by SimGC achieves even better results than the whole graph on Cora and Citeseer. This can be attributed to the fact that the condensed graph effectively captures the essential aspects of the original graph while disregarding distracting information. 

\noindent$\blacktriangleright$ \textbf{Observation 3.} SimGC achieves rather promising results even with extremely small reduction rates. For instance, the Reddit dataset can be reduced to 0.05$\%$, 0.1\% and 0.2$\%$ while still maintaining 97.0$\%$, 98.0\% and 98.6$\%$ GNN performance, respectively. The Ogbn-arxiv dataset can be compressed to 0.05$\%$, 0.25\% and 0.5$\%$ while maintaining 89.1$\%$, 93.0$\%$ and 93.6$\%$ GNN performance.
% \noindent$\blacktriangleright$ \textbf{Observation 4.} As the reduction rate escalates, the performance of SimGC exhibits an improving trend, approaching the performance of the entire dataset. This intuitively makes sense as higher reduction rates generally suggest greater fidelity. It implies that more information gets conserved in the compressed graph, aiding in performance enhancement.

% \noindent$\blacktriangleright$ \textbf{Observation 5.} Throughout the training process, SimGC consistently demonstrates the fastest convergence speed in terms of loss and achieves the lowest training loss. 
% Moreover, it can be deduced from the figure that the inclusion of complex external parameters in GCOND and GCDM not only slows down the condensation process but also has a detrimental effect on the optimization of the condensed graph. Consequently, the condensed graphs produced by these methods exhibit lower performance in comparison.

% \begin{figure*}[!t]
% \includegraphics[width=1.0\linewidth]{relative_time.pdf}
% \caption{Performance over training time. Our method consistently achieves the fastest training speed while delivering the most promising performance.}
% \label{relative_time}
% \end{figure*}
\begin{table*}[t]
\scriptsize
\setlength{\tabcolsep}{0.7pt}
\renewcommand{\arraystretch}{0.8}
\centering
\caption{The condensation time (seconds) of proposed SimGC and graph condensation baselines on four largest datasets, \wy{where ``Pre" and ``Cond" are pre-train time and condensation time.} 
% ``d" stands for day, ``Pre" refers to pre-train duration and ``Cond" is shorthand for the condensation duration. 
``Acc Rank" means the average test accuracy rank among all baselines.}
\label{tab:totaltime}
\begin{tabular}{@{}ccrrr rrr rrr rrr c@{}}
\toprule
& & \multicolumn{3}{c}{Ogbn-arxiv (169K/1M)} 
& \multicolumn{3}{c}{Ogbn-products (2M/61M)}
& \multicolumn{3}{c}{Reddit (232K/114M)}  
& \multicolumn{3}{c}{Reddit2 (232K/23M)}
&
\multicolumn{1}{c}{Acc}\\
\cmidrule(l{2pt}r{2pt}){3-5} \cmidrule(l{2pt}r{2pt}){6-8} \cmidrule(l{2pt}r{2pt}){9-11}
\cmidrule(l{2pt}r{2pt}){12-14}
& & 0.05\% & 0.25\% & 0.5\% & 0.02\% & 0.04\% & 0.08\% & 0.05\% & 0.1\% & 0.2\% & 0.05\% & 0.1\% & 0.2\%
& \multicolumn{1}{c}{Rank}
\\
\midrule
GCOND & Total  & 13,224 & 14,292 & 18,885 & 20,092 & 25,444 & 25,818 & 15,816 & 16,320 & 19,225  & 10,228 & 10,338 & 11,138
& \multicolumn{1}{c}{3.8} \\ \midrule
GCDM & Total & 1,544 & 5,413 & 13,602 & 11,022 & 12,316 & 13,292 & 1,912 & 2,372 & 4,670  & 1,615 & 1,833 & 4,574  
& \multicolumn{1}{c}{4.4} \\ \midrule
SFGC & Total &   64,260 &   67,873 &  70,128 &   128,904 &   130,739 &   132,606 &   159,206 &    160,190 &   161,044 &   124,774 &   125,331 &  126,071  & \multicolumn{1}{c}{2.2}\\ \midrule
SGDD & Total &   15,703 &   17,915 &   21,736 &   28,528 &  39,579 &   59,622 &   46,096 &   54,165 &   55,274 &   35,304 &   38,233 &   40,987 
& \multicolumn{1}{c}{2.7} \\ \midrule
\multirow{4}{*}{\begin{tabular}[c]{@{}c@{}} SimGC \end{tabular}} 
& Pre &106 &106 &106 &500 &500 &500 &185 &185 &185 &172 &172 &172 & \multirow{4}{*}{1.7}\\ 
& Cond & 237 & 795  & 1,188 & 1,415 & 1,480 & 1,645 & 755 & 830 & 1,070 & 598 & 768 & 895 \\ 
& Total & \pmb{343}  & \pmb{901}  & \pmb{1,294}  & \pmb{1,915} & \pmb{1,980} & \pmb{2,145} & \pmb{940} & \pmb{1,115} & \pmb{1,176} & \pmb{770} & \pmb{940} & \pmb{1,067}\\
& Speedup   & \pmb{4.5x}  & \pmb{6.0x}  & \pmb{10.5x} & \pmb{5.8x} & \pmb{6.2x} & \pmb{6.2x} & \pmb{2.0x} & \pmb{2.1x} & \pmb{4.0x} & \pmb{2.1x} & \pmb{2.0x} & \pmb{4.3x}
\\ 
\bottomrule
\end{tabular}
\end{table*}
\subsection{Condensation Time}
In this section, we present a comparison of the condensation time between SimGC and four condensation baselines. To ensure a fair assessment, we report both the graph condensation time and the SGC pre-training time of SimGC. By summing these times, we calculate the total time and compare it with other methods.
Based on Table~\ref{tab:totaltime}, the pre-training time is consistently much shorter than the condensation time, which demonstrates that it is feasible and affordable to pre-train an SGC for graph condensation. When considering the total time, SimGC stands out with a significantly reduced condensation time compared to existing graph condensation methods. The computation of second derivatives to GNN parameters in GCOND and SGDD requires an impractical amount of time. SFGC, on the other hand, necessitates hundreds of teacher GNNs to achieve high performance, which is both impractical and burdensome. In comparison, SimGC condenses the graph without any external parameters, achieving a minimum of two times and up to 10 times acceleration in the condensation process.

\subsection{Generalizability Capability}
In this section, we illustrate the generalizability of different condensation methods using different GNN models following the condensation process. The models employed in our experiments include MLP, GCN, SGC, GraphSAGE~\cite{hamilton2017inductive}, GIN~\cite{xu2018powerful} and JKNet~\cite{xu2018representation}.
Based on the results presented in Table~\ref{tab:models}, it can be observed that the condensed graph generated by SimGC demonstrates strong generalization capabilities across various GNN architectures and datasets. 
The condensed graph produced by SimGC outperforms the other methods by a substantial margin across the vast majority of datasets and models. Notably, the average performance of SimGC surpasses that of the other methods across all datasets.
In particular, GraphSAGE tends to perform worse than other GNN models because the aggregation process of GraphSAGE has a different root weight than normal normalization, causing disruptions in node distribution and leading to inferior performance.
Additionally, MLP is consistently inferior to all methods on all datasets, suggesting that the edges in the condensed graph play a crucial role that cannot be disregarded.
% It is also worth noting that the relatively poor generalization performance of the condensed graph on the Ogbn-products dataset may be attributed to the small proportion of the training set used and the extremely small reduction rate employed.
These results provide compelling evidence for the effectiveness of SimGC in producing condensed graphs with promising generalization abilities. 

\begin{table}[t]
\scriptsize
\centering
\renewcommand{\arraystretch}{0.7}
\caption{The generalization capabilities of SimGC on four largest datasets, \wy{where ``SAGE" and ``Avg." refer to ``GraphSAGE" and the average accuracy (\%) of five GNNs.} 
}
\begin{tabu}{@{}c@{\hspace{10pt}}c@{\hspace{10pt}}c@{\hspace{10pt}} c@{\hspace{10pt}} c@{\hspace{10pt}}c@{\hspace{10pt}}c@{\hspace{10pt}}c@{\hspace{10pt}}c @{\hspace{10pt}}c@{}}
\toprule
\multicolumn{1}{l}{}                                                 & Methods      & MLP    & GCN    & SGC  & SAGE  & GIN    & JKNet     & Avg.   \\ \midrule
\multirow{5}{*}{\begin{tabular}[c]{@{}c@{}}\textsf{Ogbn-arxiv}\\(0.5$\%$)  \end{tabular}}    
&GCOND     & 43.8 & 64.0 & 63.6 & 55.9 & 60.1 & 61.6 & 61.0 \\
&GCDM     & 41.8 & 61.7 & 60.1 & 53.0 & 58.4 & 57.2 & 58.1 \\
&SFGC     & 46.6 & \pmb{66.8} & 63.8 & \pmb{63.8} & 61.9 & \pmb{65.7} & 64.4 \\
&SGDD     & 36.9 & 65.6 & 62.2 & 53.9 & 59.1 & 60.1 & 60.2 \\
&SimGC     & \pmb{44.7} & 66.3 & \pmb{{65.1}} & {62.6} & \pmb{{63.8}} & {65.4} & \pmb{{64.6}} \\\midrule
\multirow{5}{*}{\begin{tabular}[c]{@{}c@{}}\textsf{Ogbn-products}\\(0.04$\%$)  \end{tabular}}    
&GCOND     & 33.9 & 56.4 & 52.3 & 44.5 & 50.5 & 46.3 & 50.0 \\
&GCDM     & 37.1 & 54.4 & 49.0 & 48.1 & 50.4 & 49.3  & 50.2 \\
&SFGC     & 40.9 & 64.2 & 60.4 & 60.4 & 58.9 & 61.6  & 58.3 \\
&SGDD     & 25.5 & 57.0 & 50.1 & 51.5 & 51.3 & 49.5  & 51.9 \\
&SimGC     & \pmb{45.1} & \pmb{64.8} & \pmb{{60.6}} & \pmb{60.1}& \pmb{{57.6}} & \pmb{{57.8}}  & \pmb{{60.2}} \\\midrule
\multirow{5}{*}{\begin{tabular}[c]{@{}c@{}}\textsf{Reddit}\\(0.2$\%$)  \end{tabular}}    
&GCOND     & \pmb{48.4} & 91.7 & 92.2 & 73.0 & 83.6 & 87.3 & 85.6 \\
&GCDM     & 40.5 & 83.3 & 79.9 & 55.0 & 78.8 & 77.3 & 74.9 \\
&SFGC     & 45.4 & 87.8 & 87.6 & \pmb{84.5} & 80.3 & 88.2  & 85.7 \\
&SGDD     & 24.8 & 89.8 & 87.5 & 73.7 & 85.2 & 88.9  & 85.0 \\
&SimGC     & 44.6 & \pmb{{92.6}} & \pmb{{92.3}} & {84.4} & \pmb{{90.7}} & \pmb{{91.3}} & \pmb{{90.3}} \\\midrule
\multirow{5}{*}{\begin{tabular}[c]{@{}c@{}}\textsf{Reddit2}\\(0.2$\%$)  \end{tabular}}    
&GCOND     & 35.7 & 82.4 & 77.1 & 59.8 & 79.6 & 73.0 & 74.4 \\
&GCDM     & 32.5 & 84.7 & 78.0 & 55.3 & 75.3 & 70.6 & 72.8 \\
&SFGC     & 42.6 & 88.0 & 86.8 & 77.9 & 74.8 & 86.0 & 82.7 \\
&SGDD     & 25.1 & 86.0 & 87.5 & 73.1 & 79.4 & 84.7 & 82.1 \\
&SimGC     & \pmb{44.8} & \pmb{{91.3}} & \pmb{{91.4}} & \pmb{{81.9}}
& \pmb{{88.9}} & \pmb{{89.0}} & \pmb{{88.5}} \\ \bottomrule
\end{tabu}
\label{tab:models}
\end{table}

\subsection{Neural Architecture Search}
In this section, our objective is to evaluate the effectiveness of our approach in conducting NAS. We perform a NAS experiment on the condensed graph and compare it to a direct search on the original graph. To assess the performance of the search process, we utilize two benchmark datasets: Ogbn-arxiv and Reddit. The procedure for conducting the search on the condensed graph is outlined as follows: (1) Utilize the original graph or the condensed graph to train 54 GNNs (GNN$_{S}$) by using different GNN architectures from the search space. (2) Determine the optimal GNN architecture according to the best-performing GNN$_{S}$ evaluated on the original validation set. (3) Train the optimal GNN$_{S}$ architecture on the original graph to obtain the result GNN$_{NAS}$.
\begin{table}[t]
\scriptsize
\setlength{\tabcolsep}{1pt}
\renewcommand{\arraystretch}{0.8}
\centering
\caption{Evaluation of \wy{NAS with} test accuracy (\%) on the Ogbn-arxiv and Reddit datasets. ``BT" and ``AT" refer to evaluation before tuning and after tuning. ``WD" refers to searching with the whole dataset.}
\begin{tabu}{@{}c@{\hspace{10pt}}c@{\hspace{10pt}}c@{\hspace{10pt}}c@{\hspace{10pt}}c@{\hspace{10pt}}c@{\hspace{10pt}}c@{}}
\toprule
\multirow{2}{*}{}  & \multicolumn{2}{c}{GNN$_{S}$}  & \multicolumn{2}{c}{GNN$_{N\!A\!S}$} & \multicolumn{2}{c}{Time}\\ \cmidrule(l){2-3} \cmidrule(l){4-5}  \cmidrule(l){6-7} 
&   BT &   AT & SimGC & WD    & SimGC & WD\\ \midrule
\multicolumn{1}{c}{\begin{tabular}[c]{@{\hspace{10pt}}c@{\hspace{10pt}}}Ogbn-arxiv  (0.5\%) \end{tabular}}
& 66.5    & 67.8   & 71.3  & 72.1  & 16s & 104s \\   \midrule
\multicolumn{1}{c}{\begin{tabular}[c]{@{\hspace{10pt}}c@{\hspace{10pt}}}Reddit  (0.2\%)   \end{tabular}} & 92.6    & 93.7   & 94.3  & 94.6  & 74s & 518s    \\    
\bottomrule
\end{tabu}
\label{tab:nas}
\end{table}
\begin{table}[t]
\scriptsize
\setlength{\tabcolsep}{8pt}
\renewcommand{\arraystretch}{0.9}
\centering
\caption{Knowledge Distillation with Condensed Graphs. ``Whole" denotes the use of the original graph for knowledge distillation. ``w/o KD" means \wy{directly} training on the condensed graph without knowledge distillation. ``with KD" means training by our knowledge distillation. 
% ``Params" refers to the total number of parameters, ``Inference" indicates the time taken to perform inference on the entire dataset, and ``Training" denotes the time utilized for the whole KD process. 
}
\begin{tabu}{@{}lcccc@{}}
\toprule
Models & Params & Inference & Training & Accuracy \\ \midrule
\multirow{1}{*}{\begin{tabular}[c]{@{}c@{}}\textsf{Teacher}  \end{tabular}}    
& 348,160 & 0.396s & 421s & 71.4\% \\\midrule
\multirow{1}{*}{\begin{tabular}[c]{@{}c@{}}\textsf{Student 1 (Whole)} \end{tabular}}    
& 10,752 & 0.036s & 218s & 69.5\% \\
\multirow{1}{*}{\begin{tabular}[c]{@{}c@{}}\textsf{Student 1 (w/o KD)} \end{tabular}}    
& 10,752 & 0.042s & 24s & 64.7\% \\
\multirow{1}{*}{\begin{tabular}[c]{@{}c@{}}\textsf{Student 1 (with KD)} \end{tabular}}    
& 10,752 & 0.037s & 27s & 65.4\% \\\midrule
\multirow{1}{*}{\begin{tabular}[c]{@{}c@{}}\textsf{Student 2 (Whole)}   \end{tabular}}    
& 43,008 & 0.056s & 298s & 70.7\% \\
\multirow{1}{*}{\begin{tabular}[c]{@{}c@{}}\textsf{Student 2 (w/o KD)}   \end{tabular}}    
& 43,008 & 0.044s & 26s & 66.5\% \\
\multirow{1}{*}{\begin{tabular}[c]{@{}c@{}}\textsf{Student 2 (with KD)}  \end{tabular}}    
& 43,008 & 0.044s & 29s & 67.4\% \\ \bottomrule
\end{tabu}
\label{tab:kd}
\end{table}

By comparing the performance of GNN$_{NAS}$ obtained from both the condensed graph and the original graph, we can evaluate the effectiveness of the condensed graph approach in guiding the NAS process. The comprehensive search space encompasses a wide range of combinations involving layers, hidden units, and activation functions:
\begin{itemize} % [leftmargin=*]
\item[$\bullet$] Number of layers are searched over \{2, 3, 4\}.
\item[$\bullet$] Hidden dimension are searched over \{128, 256, 512\}.
\item[$\bullet$] \{$\operatorname{Sigmod}(\cdot)$, $\operatorname{Tanh}(\cdot)$, $\operatorname{Relu}(\cdot)$, $\operatorname{Softplus}(\cdot)$, $\operatorname{Leakyrelu}(\cdot)$, $\operatorname{Elu}(\cdot)$\} are activation functions.
\end{itemize}
% \begin{itemize}
% [nosep,leftmargin=1em,labelwidth=*,align=left]
% \item[\bullet] Number of layers are searched over \{2, 3, 4\}.
% \item[\bullet] Hidden dimension are searched over \{128, 256, 512\}.
% \item[\bullet] $\operatorname{sigmod}(\cdot)$, $\operatorname{tanh}(\cdot)$, $\operatorname{relu}(\cdot)$, $\operatorname{softplus}(\cdot)$, $\operatorname{leakyrelu}(\cdot)$, $\operatorname{elu}(\cdot)$ are activation functions.
% \end{itemize}
Table~\ref{tab:nas} presents our research findings including (1) the test accuracy of GNN$_{S}$ before and after tuning, (2) the NAS result GNN$_{N\!A\!S}$ and (3) the average time of searching a single architecture on both the condensed and original graphs. Based on our experimental results, we observe that SimGC's condensed graph exhibits the ability to efficiently tune parameters and improve performance. Specifically, the condensed graphs of both datasets demonstrate an increase in performance by over 1\%. Furthermore, we observe that the results of NAS using the condensed graph are comparable to that of using the original graph, while the search time for the condensed graph is over six times faster. These findings suggest that utilizing the condensed graph for NAS is a viable option.

\begin{table*}[t]
\tiny
\centering
\renewcommand{\arraystretch}{0.9}
\caption{Statistics comparison between condensed graphs and original graphs.}
\label{tab:con vs ori}
\setlength{\tabcolsep}{2pt}
\begin{tabular}{@{}l cc cc cc cc cc cc cc@{}}
\toprule
& \multicolumn{2}{c}{Citeseer, $r$=1.8\%} & \multicolumn{2}{c}{Cora, $r$=2.6\%} & \multicolumn{2}{c}{Ogbn-arxiv, $r$=0.25\%}
& \multicolumn{2}{c}{Ogbn-products, $r$=0.04\%}
& \multicolumn{2}{c}{Flickr, $r$=0.5\%} 
& \multicolumn{2}{c}{Reddit, $r$=0.1\%} 
& \multicolumn{2}{c}{Reddit2, $r$=0.1\%}
\\ %\midrule
\cmidrule{2-3} \cmidrule{4-5} \cmidrule{6-7} \cmidrule{8-9} \cmidrule{10-11} \cmidrule{12-13} \cmidrule{14-15}
& Whole              & SimGC            & Whole            & SimGC          & Whole                 & SimGC            & Whole             & SimGC           & Whole               & SimGC        
& Whole              & SimGC
& Whole              & SimGC\\ \midrule
Accuracy  & 71.7\%               & 72.2\%               & 81.2\%             & 80.9\%             & 71.4\%                  & 66.4\%
& 74.0\%                  & 66.1\%
& 47.2\%              & 45.6\%              
& 93.9\%                & 92.0\%            
& 93.5\%                & 91.0\%
\\
Nodes   & 3,327              & 60                 & 2,708            & 70               & 169,343               & 477
& 2,449,029               & 1,010
& 44,625            & 227               
& 153,932             & 177             
& 153,932             & 177\\
Edges   & 4,732              & 1,414               & 10,556            & 754             & 1,166,243             & 4,534
& 61,859,140             & 15,346
& 218,140           & 1,393              
& 10,753,238          & 2,865
& 10,753,238          & 4,113
\\
Sparsity  & 0.09\%              & 39.28\%             & 0.15\%            & 15.47\%           & 0.01\%                 & 1.99\% 
& 0.001\%                 & 1.50\%
& 0.02\%             & 2.22\%            
& 0.09\%               & 9.14\%
& 0.09\%               & 13.13\%\\
Storage   & 47.1 MB            & 0.9 MB             & 14.9 MB          & 0.4 MB           & 100.4 MB              & 0.3 MB
& 284.1 MB              & 0.7 MB
& 86.8 MB           & 0.5 MB            & 435.5 MB            & 0.5 MB          
& 435.5 MB              & 0.5 MB
\\ \bottomrule
\end{tabular}
\end{table*}

\begin{figure}[t]
\centering
\includegraphics[width=0.7\textwidth]{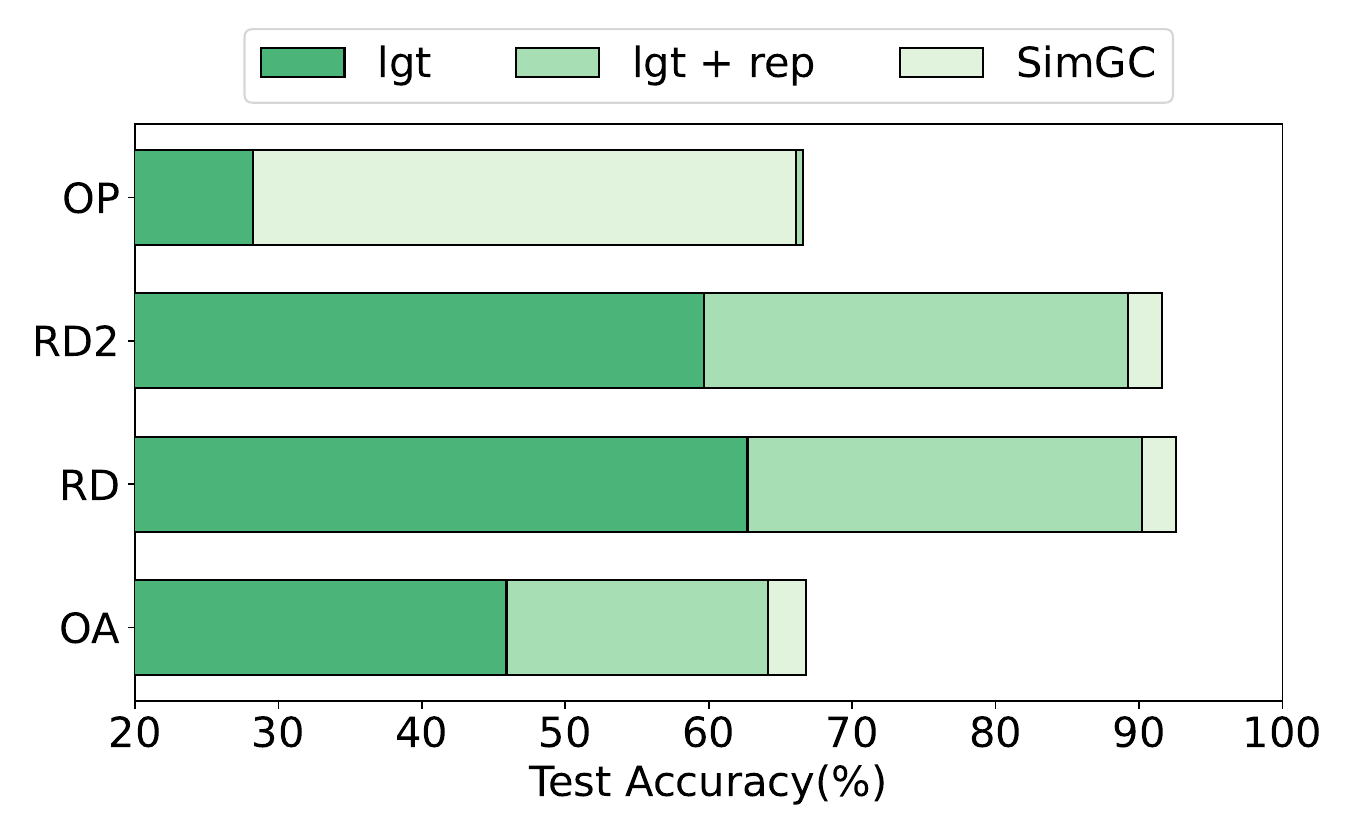}
\caption{Comparison of the proposed SimGC with two variants, \wy{where} Ogbn-arxiv, Reddit, Reddit2 and Ogbn-products are denoted as ``OA", ``RD", ``RD2", ``OP". 
% ``w/o" means ``without".
}
\label{fig:ablation}
\end{figure}

\subsection{Knowledge Distillation}
One notable advantage of the condensed graph is its capability to leverage a pre-trained GNN (SGC) for lightweight knowledge distillation. In this section, we will delve into the potential of the condensed graph to facilitate knowledge distillation.
We conduct knowledge distillation experiments on the condensed graph of Ogbn-arxiv (0.5\%). The procedure is as follows: Initially, both the teacher and student GNNs create predictions on the condensed graph to obtain output logits. Subsequently, the KL divergence of the two logits acts as the soft loss, while the condensed graph's classification loss serves as the hard loss, both of which are employed to optimize the student model. The architectures of the teacher and student models are presented as follows:
\begin{itemize} % [leftmargin=*]
\item[$\bullet$] \textbf{Teacher}: A three-layer GCN with hidden features of \{512, 512, 40\}.
\item[$\bullet$] \textbf{Student 1}: A two-layer GCN with hidden features of \{64, 40\}.
\item[$\bullet$] \textbf{Student 2}: A two-layer GCN with hidden features of \{256, 40\}.
\end{itemize}
The results presented in Table~\ref{tab:kd} illustrate that the utilization of knowledge distillation can enhance the performance of the GNN trained on the condensed graph, while significantly reducing the training time compared to direct distillation with the original graph. However, it is important to note that the performance achieved through knowledge distillation using the condensed graph is still noticeably lower than the performance obtained through direct knowledge distillation with the original graph. This result is reasonable considering the extremely small size of the condensed graph. Nevertheless, the use of condensed graphs for knowledge distillation holds the potential for reducing time and GPU memory requirements in large-scale graphs, making it feasible to perform knowledge distillation on large-scale graphs even in environments with limited resources.

\subsection{Statistics of Condensed Graphs}
In Table~\ref{tab:con vs ori}, we compare several properties between condensed graphs and original graphs and make several observations. Firstly, the condensed graphs have a smaller number of nodes and edges, leading to significantly reduced storage requirements while achieving similar or even higher performance for downstream tasks.
Secondly, the condensed graphs exhibit lower sparsity compared to their larger counterparts. 
% This is primarily due to the condensed graph being on an extremely small scale. If the original sparsity were maintained, it would result in minimal connections between nodes in the condensed graph.

\subsection{Ablation Study}
% In this subsection, we evaluate the effectiveness of the representation alignment and smoothness regularizer proposed in the previous subsection. 
\xzb{In this subsection, we compare SimGC with two variants of our method: SimGC with only logit alignment (without representation alignment and feature smoothness), SimGC with logit and representation alignment (without feature smoothness).} We perform the experiment on the four largest datasets: Ogbn-arxiv, Reddit, Reddit2 and Ogbn-products.
The results are summarized in Fig~\ref{fig:ablation}. Initially, when considering logit alignment alone, SimGC demonstrates poor performance. When incorporating the representation alignment together, we observe a significant improvement of over 20\% across all datasets, demonstrating its ability to explicitly align the node feature and adjacency matrix. Notably, the most promising outcomes are usually achieved when the representation alignment, logit representation alignment and feature smoothness regularizer are employed simultaneously, demonstrating the effectiveness of our framework over the other variants.
Furthermore, we notice that the feature smoothness regularizer performs better on datasets with high homophily (Reddit and Reddit2) and degrades on datasets with low homophily (Ogbn-products).
% Furthermore, the feature smoothness regularizer exhibits superior performance in datasets characterized by higher homophily, such as Reddit and Reddit2, but may cause performance decrease in datasets with low homophily, like Ogbn-products. This suggests that we can consider using the feature smoothness regularizer depending on the dataset's homophily.

\section{Conclusion}
This paper introduces the Simple Graph Condensation (SimGC) framework, which offers a straightforward yet powerful approach to simplify metric alignment in graph condensation, aiming to reduce unnecessary complexity inherited from intricate metrics. SimGC removes external parameters and focuses solely on aligning the condensed graph with the original graph, from the input layer to the prediction layer, guided by a pre-trained SGC model on the original graph.
Extensive experimental results demonstrate the superiority of SimGC in terms of prediction accuracy, condensation time and generalizability. 
For future work, we intend to expand the framework to accommodate various types of graphs, including heterogeneous graphs and hypergraphs, making SimGC a universally applicable framework with broader applicability.

\section{Acknowledgements}
This research was supported by the Joint Funds of the Zhejiang Provincial Natural Science Foundation of China (No. LHZSD24F020001), Zhejiang Province "JianBingLingYan+X" Research and Development Plan (No. 2024C01114), Ningbo Natural Science Foundation (No. 2023J281), and Zhejiang Province High-Level Talents Special Support Program "Leading Talent of Technological Innovation of Ten-Thousands Talents Program" (No. 2022R52046).


% \clearpage
% \bibliography{ref}

\end{document}